\documentclass[cameraready]{Interspeech}

\newcommand{\ourmethod}{MultiLinguahah }

\title{\ourmethod: A New Unsupervised Multilingual Acoustic Laughter Segmentation Method}

\author[affiliation={1,2}]{Sofia}{Callejas}
\author[affiliation={2}]{Nahuel}{Gomez}
\author[affiliation={3}]{Catherine}{Pelachaud}
\author[affiliation={*1}]{Brian}{Ravenet}
\author[affiliation={*2}]{Valentin}{Barriere}

\address{
    $^1$ Université Paris-Saclay LISN -- Orsay, France \\
    $^2$ Universidad de Chile DCC -- Santiago, Chile \\
    $^3$ Sorbonne University ISIR -- Paris, France
}

\email{@universite-paris-saclay.fr, @dcc.uchile.cl, @sorbonne-universite.fr}

\keywords{laughter, audio segmentation, computational paralinguistics}

\usepackage{comment}
\usepackage{multirow}
\usepackage{tikz}
\usepackage{graphicx}
\usetikzlibrary{shapes.geometric, arrows.meta, positioning, calc}

\usepackage{multicol}
\usepackage{booktabs}

\begin{document}

\maketitle
\def\thefootnote{*}\footnotetext{Same supervision}
\def\thefootnote{\arabic{footnote}} 

\begin{abstract}

Laughter is a social non-vocalization that is universal across cultures and languages, and is crucial for human communication, including social bonding and communication signaling.
However, detecting laughter in audio is a challenging task, and segmenting it is even more difficult. Currently, Machine Learning methods generally rely on costly manual annotation, and their datasets are mostly based on English contexts.
Thus, we propose an unsupervised multilingual method that sets up the laughter segmentation task as an anomaly detection of energy-based segmented audio sequences. Our method applies an Isolation Forest on audio representations learned from BYOL-A encoder. 
We compare our method with several state-of-the-art laughter detection algorithms on four datasets, including stand-up comedy, sitcoms, and general short audio from AudioSet.
Our results show that state-of-the-art methods are not optimized for multilingual contexts, while our method outperforms them in non-English settings.

\end{abstract}

\section{Introduction}

Laughter is ever-present in human interactions, playing an important part in human-human communication, acting also as a tool for social bonding \cite{Dunbar2012}\cite{Provine2006}. 
It is inherently social, as it not only communicates one's internal state but also helps to propagate this state to other listeners \cite{Strick2021}. 
It can express joy, relief, or success, but also appears during embarrassment, anger, or even sadness \cite{glenn2003laughter}. 


Laughter has been the object of study in domains such as psychology \cite{wood2022semantic}, linguistics \cite{ginzburg2020laughter}, or 
computer science \cite{dupont2016laughter,Hyun2024}. 
%
Laughter is useful for a broad variety of different tasks crucial to understand human communication and for building Socially Interactive Agents (SIAs; \cite{lugrin2022handbook}): from emotion recognition to humor extraction \cite{Hasan2019b,Hyun2024} and eventually natural speech generation \cite{Xin2023}. This makes its automatic detection important. 

One of the perspectives from which laughter can be studied is its acoustic structure. Recent research suggests that this structure is universal, meaning that laughter exhibits similar acoustic patterns across languages \cite{bryant2022laughter}, impulsing the need for general language and domain-agnostic systems detecting it.

Deep Learning (DL) algorithms are robust methods for automatic laughter detection and laughter segmentation, where they aim at detecting the beginning and end of laughter in an audio file. But the most efficient algorithms to tackle this task are DL-based
\cite{Omine2024,Gillick2021,Matsuda2023} and generally require a high quantity of manual annotation. 
Requiring the exact timestamps is a meticulous task that is very time-consuming, making unsupervised modelling very useful in this case.

Recent works, when not relying on manual annotations, proposed to annotate laughter using automatic methods such as using the transcripts \cite{Hasan2019b}, or artificially adding laughter to non-laughter segments \cite{Omine2024}. The former does not give the exact position, while the latter allows for applying data-augmentation techniques right on the laughter and knowing precisely where the exact laughter starts and ends in the new audio. However, naturally occurring examples are more diverse than artificially created data, as they can drift from the original data distribution \cite{Naous2024,quiroga-etal-2025-adapting,Barriere2024}. 
For instance, unsupervised methods such as FunnyNet \cite{Liu2024b} are evaluated on datasets like \textit{Friends}, a sitcom in which speech is dominant, background noise is limited, and the sound mixing is professionally controlled. 
However, this setting does not reflect real-world multilingual data collected from diverse countries and acoustic environments. In such data, background music, environmental noise, and recording conditions vary considerably. 
In-the-wild stand-up comedy offers the type of diverse data containing challenges that can face acoustic laughter segmentation systems \cite{Ryokai2018}: diverse audio capture conditions, as well as laughter types and languages. We manually annotate test data in various languages to validate our results. 

In this paper, we propose an unsupervised acoustic laughter segmentation model, which performs well for different domains and languages. We show that state-of-the-art methods, if generally better on US English, are not consistently performant across languages and domains. 
Overall, the main contributions of this paper can be summarized as follows: \textit{(i)} a new unsupervised method for laughter segmentation based on energy-based segmentation and isolation forests, \textit{(ii)} applied across languages and domains through publicly available annotated datasets and new laughter annotations, \textit{(iii)} analysed with respect to the laughter duration. Our code is available online: \url{https://github.com/sofia-callejas/Multilinguahah}.





\section{\ourmethod: Acoustic Laughter Segmentation}

\begin{figure*}
    \centering
    \begin{tikzpicture}[
    node distance=0.5cm and 0.7cm,
    block/.style={rectangle, draw, fill=white, text width=4.8cm, align=center, minimum height=1cm, font=\tiny\sffamily, rounded corners=1pt},
    iso/.style={rectangle, draw, fill=white, text width=4.5cm, align=center, minimum height=3.5cm, font=\tiny\sffamily, rounded corners=1pt},
    voice/.style={rectangle, draw, fill=white, text width=4.8cm, align=center, minimum height=1.2cm, font=\tiny\sffamily, rounded corners=1pt},
    stepblock/.style={rectangle, draw, fill=white, text width=3cm, align=center, minimum height=0.5cm, font=\tiny\sffamily},
    encoderblock/.style={trapezium, draw, fill=blue!10, trapezium left angle=75, trapezium right angle=75, text width=1.2cm, align=center, minimum height=1.5cm, font=\tiny\bfseries\sffamily, shape border rotate=90},
    arrow/.style={-Stealth, thin},
]

\node (input) [block] {
    \includegraphics[width=4.8cm]{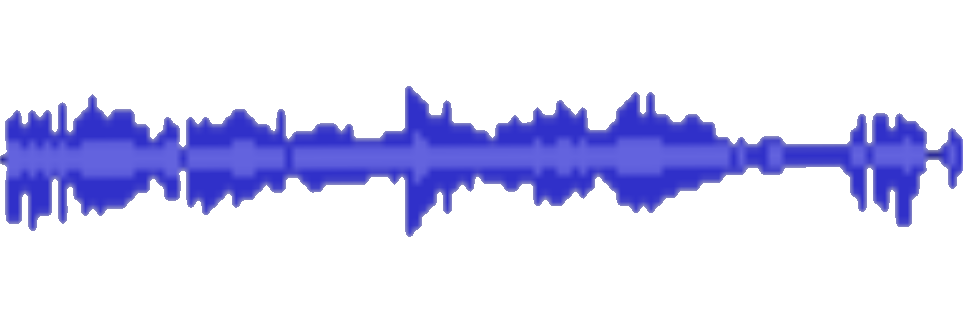} \\ 
    \textbf{Audio Input} 
};

\node (voice_rem) [voice, below=of input] {
    \includegraphics[width=4.8cm]{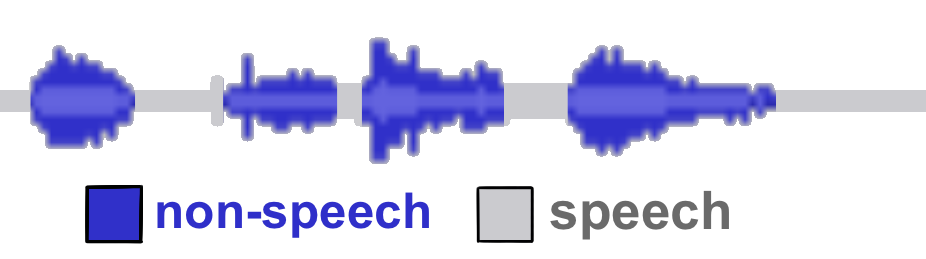} \\ 
    \S \ref{subsec:voice_removal} \textbf{Voice Removal} 
};

\node (step_thresh) [stepblock, above right=0cm and 0.5cm of voice_rem] {
    \includegraphics[width=3cm]{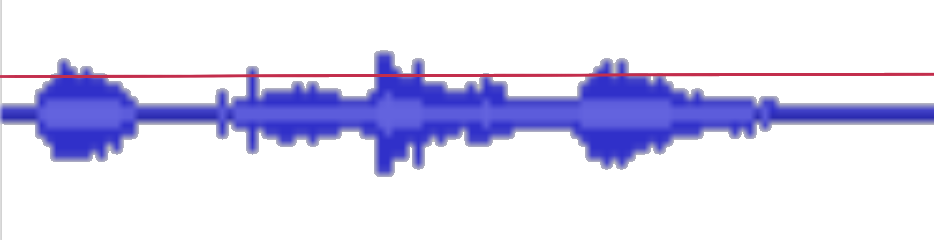} \\ 
    \S \ref{subsec:energy_segmentation} \textbf{Energy Threshold}
};

\node (encoder) [encoderblock, right=of step_thresh] {
    \S \ref{subsec:audio_encoding} Encoder \\  $\mathcal{E}$
};

\node (iforest) [iso, right= of encoder] { \S \ref{subsec:anomaly_detection}
    \textbf{Isolation Forest}  $\mathcal{A}$ \\ \includegraphics[width=4.5cm]{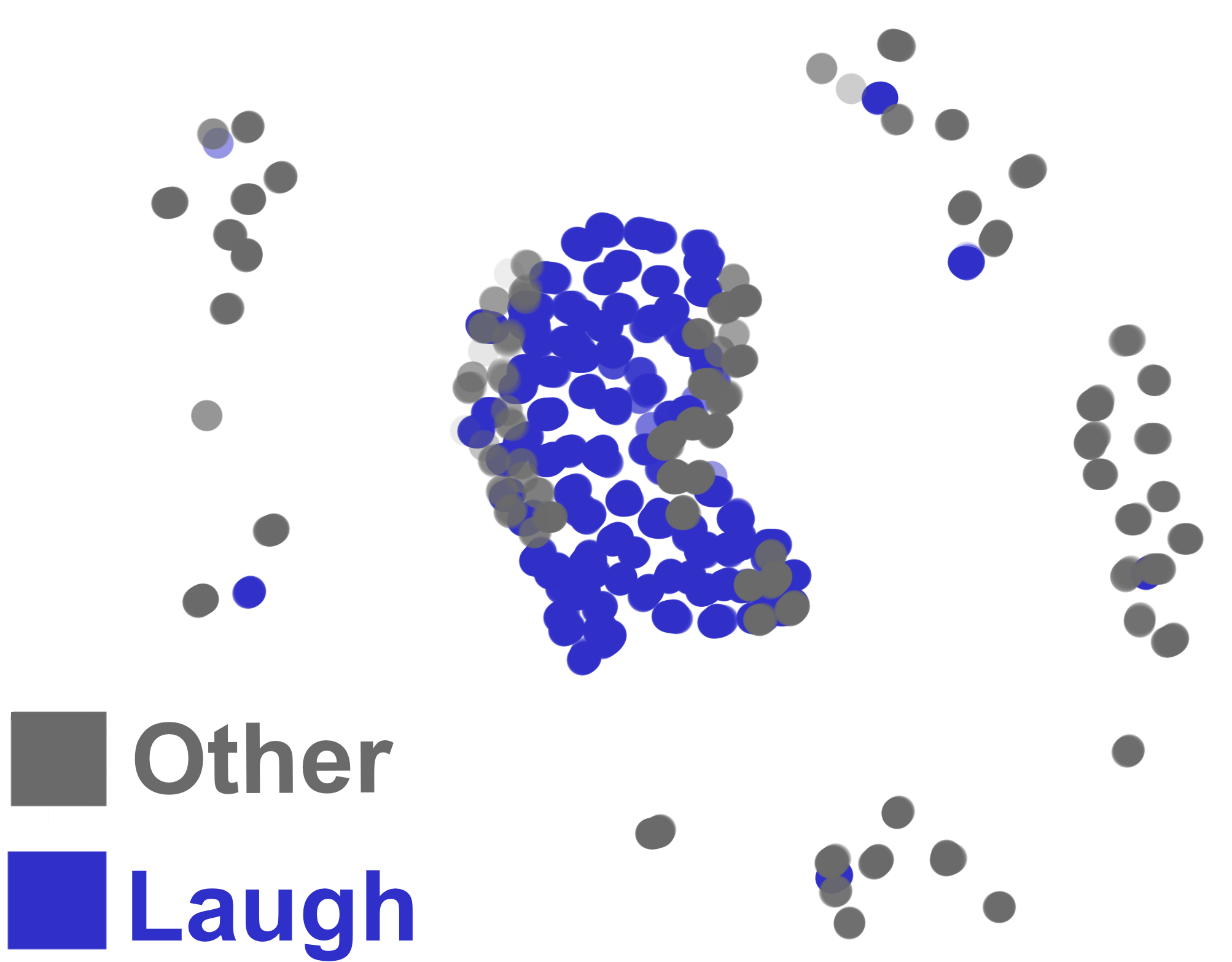} \\ \tiny(Anomaly Detection)
};

\draw [arrow] (input) -- (voice_rem);
\draw [arrow] (voice_rem) -- (step_thresh);
\draw [arrow] (step_thresh) -- (encoder)  ; 
\draw [arrow] (encoder) -- (iforest);

\draw[dashed, gray] ($(step_thresh.north west)+(-0.05,0.05)$ ) rectangle ($(step_thresh.south east)+(0.05,-0.05)$);

\end{tikzpicture}
    
    \caption{
    We first remove the voices from the laughter through channel subtraction or audio source separation (\S \ref{subsec:voice_removal}), second then we segment the audio into events using an energy-based threshold (\S \ref{subsec:energy_segmentation}), third we encode the audio using a pre-trained model (\S \ref{subsec:audio_encoding}), and finally we detect laughter using an anomaly detection algorithm based on Isolation Forest (\S \ref{subsec:anomaly_detection}).}
    \label{fig:overview}
\end{figure*}
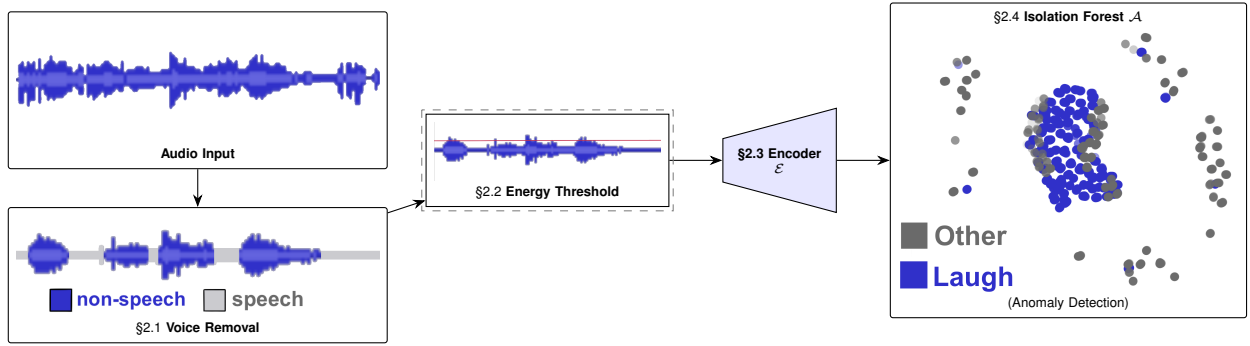

The proposed method is composed of several steps. An overview is shown in Figure \ref{fig:overview}. 



\subsection{Voice Removal} \label{subsec:voice_removal}

The first step of our approach consists of removing the speech from the audio signal, in order to retain the background, including laughter, music, and environmental sounds. 

In order to isolate the human voices from the rest of the audio, we apply an off-the-shelf Audio Source Separation  model $\mathcal{S}$ to the raw audio. This model allows for the separation of speech signals from non-vocal acoustic interference. 
In particular, we used a basic deep learning model based on a densely connected Convolutional Neural Network architecture \cite{Huang2017} specifically designed for vocal signal separation \cite{Takahashi2017}.

In TV shows that have studio-recorded audio, laughter comes from the crowd and not from the interactants themselves. It is possible to focus only on the non-voice audio by subtracting the two audio channels. This technique was used on the Friend dataset, following the protocol of \cite{Liu2024b}. 







\subsection{Energy-based Audio Segmentation} \label{subsec:energy_segmentation}

Once the voice has been separated from the other sources, the audio is now composed of several non-speech events that can be music, ambient sounds, or laughter. 
We use an energy-based peak detector \footnote{\url{https://github.com/amsehili/auditok}} to find the beginning and end times of every event. Using the waveform energy as a threshold, these events are removed, retaining only the non-speech audio. The threshold was arbitrarily chosen so that slight background noise was not included in the non-silence segments. Using a lower threshold allows for more noisy events to be included, for example, if detecting very weak laughter is interesting.  






\subsection{Audio Encoding} \label{subsec:audio_encoding}

Once the non-speech audio events have been extracted, a pre-trained encoder $\mathcal{E}$ was used to transform these events into vectors. For this purpose, we relied on BYOL-A \cite{Niizumi2023}, a self-supervised learning method for audio that learns general-purpose representations without requiring labelled data, and that has been shown useful for non-semantic speech tasks \cite{Calbucura2025}. 
This model is initialized with weights obtained from self-supervised pretrained on the balanced and unbalanced training splits of AudioSet \cite{Audioset}, comprising $1.963.807$ audio segments (aprox. $5.455$ hours), as well as the dataset FSD50K \cite{FSD50K}, which contains $40.966$ audio ($80$ hours). 
%

Furthermore, to perform domain adaptation, we incorporate the unlabelled training split of the target dataset into the self-supervised pre-training stage. 

\subsection{Anomaly Detection} \label{subsec:anomaly_detection}

Finally, in order to separate the vectors of the laughter events from the other ones, we utilize an unsupervised algorithm for anomaly detection $\mathcal{A}$. 
Isolation Forest \cite{isolationforest} model was used to isolate outliers by recursively partitioning data using random feature splits. 
Laughter shows consistent acoustic characteristics across languages, whereas background music and other noises differ and are treated as anomalies by the model.

\section{Experiments and Results}

\subsection{Datasets for Evaluation} 

We are validating and comparing models on a selection of 4 datasets containing laughter from various domains (in-the-wild, studio-recorded, and artificially created). \\

\noindent
\textbf{StandUp4AI} \cite{barriere2025standup4ai} dataset consists of 3,617 stand-up comedy videos spanning 7 languages. It includes audience laughter annotations, capturing performances from comedians in diverse linguistic and cultural contexts. To construct the dataset, the authors collected a targeted set of stand-up comedy videos from online platforms. We added new annotations to this dataset in US English, Canadian French, and Latin American Spanish. We used the test part of the dataset, composed of 100 videos, for 8.53 hours of audio and 3.453 laughter events. \\

\noindent
\textbf{AudioSet} \cite{Gemmeke2017} is a large-scale dataset of audio segments, each approximately 10 seconds long, sourced from \textbf{YouTube} videos and labelled with over 500 audio event categories. The dataset features diverse acoustic environments and recording qualities, covering a wide range of sound types, for 5.8 thousand hours. Laughter has been added to the audio sequences, making the timestamps known but the data artificial. The test part of the dataset was used, comprising 724 available videos and a total of 1,252 annotated laughter instances.\\

\noindent
\textbf{Friends} \cite{Brown2021} dataset is from the homonym sitcom, and consists of all 25 episodes from its third season, each approximately 23 minutes long, totalling around 10 hours of audio-visual content. The test set consists of the final 5 episodes (episodes 21–25). Within this test partition, laughter events have been manually annotated, identifying a total of 924 distinct instances of laughter. \\


\noindent
\textbf{Kuznetsova} \cite{Kuznetsova2024} is a bilingual dataset containing stand-up videos in English and Russian. The Russian subset includes 46 videos (17 hours) from 8 YouTube channels, mostly from a Vladivostok-based club. 
The English subset includes 56 videos (20 hours) from the largest stand-up YouTube channel. 
In our experiments, we focus on the official test set that has been manually annotated by the original authors. This test partition comprises 10 videos in total, five in Russian (RU) and five in English (US EN). The combined duration of this evaluation set is 1.18 hours, containing a total of 617 annotated laughter instances.

\subsection{Models for Evaluation}

We compare our approach with three baseline models: 

\textbf{Gillick et al.} \cite{Gillick2021} model is ResNet-based model learned in a supervised way, using classical data-augmentation audio techniques (pitch-shifting, time-stretching, and artificial reverberation) at the level of the audio. Data-augmentation allows the model to take advantage of the SwitchBoard annotated data and generalize to in-the-wild laughter detection. We follow the protocol of the authors to use the model for laughter segmentation (frame-level detection).  

\textbf{Omine et al.} \cite{Omine2024} model is a fine-tuning of wav2vec 2.0, 
trained in a supervised way on a large amount of data by randomly synthesizing data-augmented laughter in audio data of various sound qualities and from various recording environments. Using a data augmentation technique, laughter samples were inserted into the Non-laughter audio to create realistic training examples. 
Non-laughter audio was drawn from the Spotify Podcast Dataset \cite{Clifton2020} and AudioSet \cite{Gemmeke2017}, and laughter samples from VocalSound \cite{Gong2022} and Laughterscape \cite{Xin2023}. 

\textbf{Liu et al.} \cite{Liu2024b} model is an unsupervised baseline method \cite{Liu2024b}. This baseline applies a similar methodology: it subtracts the channel to remove the voice and uses an energy-based peak detector to detect events.
However, instead of employing an anomaly detection, it applies a K-means clustering algorithm to group the latent representations. For laughter detection, all clusters are retained except the smallest ones, which are assumed to correspond to non-laughter segments. With respect to the protocol, we used the same source separator as the one applied in our method when processing in-the-wild audio. 

Finally, we also compare them with a mix of our model and Omine's model, showing that they can complement each other. 

\begin{table}[!h]
\centering
\resizebox{.97\columnwidth}{!}{
\begin{tabular}{lllcc}
\toprule
Lang & Domain & Method & \multicolumn{2}{c}{F1} \\
     &        &        & IoU=0.3 & IoU=0.7 \\
\midrule

\multirow{15}{*}{US EN}
& \multirow{5}{*}{Stand-up}
& Gillick et al. \cite{Gillick2021} & 0.456 &  0.134 \\
& & Omine et al. \cite{Omine2024} & \textbf{0.679} & \textbf{0.356}  \\
& & Liu et al. \cite{Liu2024b} & 0.447 & 0.145  \\
& & \ourmethod & 0.506  & 0.176 \\
& & Omine+\ourmethod & 0.670 & 0.325  \\

\cmidrule(lr){2-5}

& \multirow{5}{*}{TV Show}
& Gillick et al. \cite{Gillick2021} & 0.646 & 0.197 \\
& & Omine et al. \cite{Omine2024} & 0.189 & 0.054 \\
& & Liu et al. \cite{Liu2024b} & 0.878  & 0.503  \\
& & \ourmethod & \textbf{0.910} & \textbf{0.735} \\
& & Omine+\ourmethod & 0.848 & 0.682 \\

\cmidrule(lr){2-5}

& \multirow{5}{*}{Youtube}
& Gillick et al. \cite{Gillick2021} & 0.544 & 0.220 \\
& & Omine et al. \cite{Omine2024} & 0.555 & \textbf{0.238} \\
& & Liu et al. \cite{Liu2024b} & 0.362 & 0.066  \\
& & \ourmethod & 0.315 & 0.087  \\
& & Omine+\ourmethod & \textbf{0.656} & 0.206  \\

\midrule \midrule

\multirow{5}{*}{UK EN}
& \multirow{5}{*}{Stand-up}
& Gillick et al. \cite{Gillick2021} & 0.565 & 0.132 \\
& & Omine et al. \cite{Omine2024} & 0.626  & 0.294  \\
& & Liu et al. \cite{Liu2024b} & 0.733  & 0.394  \\
& & \ourmethod & 0.736  & 0.398  \\
& & Omine+\ourmethod & \textbf{0.756}  & \textbf{0.403}  \\

\midrule

\multirow{5}{*}{ES}
& \multirow{5}{*}{Stand-up}
& Gillick et al. \cite{Gillick2021} & 0.294 & 0.076 \\
& & Omine et al. \cite{Omine2024} & 0.361 & 0.120  \\
& & Liu et al. \cite{Liu2024b} & 0.654 & 0.305  \\
& & \ourmethod & 0.649 & \textbf{0.306}  \\
& & Omine+\ourmethod & \textbf{0.676} & 0.303  \\

\midrule

\multirow{5}{*}{Lat. ES}
& \multirow{5}{*}{Stand-up}
& Gillick et al. \cite{Gillick2021} & 0.245 & 0.031 \\
& & Omine et al. \cite{Omine2024} & 0.332 & 0.133 \\
& & Liu et al. \cite{Liu2024b} & 0.572 & 0.187  \\
& & \ourmethod & 0.587 & 0.193 \\
& & Omine+\ourmethod & \textbf{0.609} & \textbf{0.205}  \\

\midrule

\multirow{5}{*}{FR}
& \multirow{5}{*}{Stand-up}
& Gillick et al. \cite{Gillick2021} & 0.149 & 0.009 \\
& & Omine et al. \cite{Omine2024} & 0.257 & 0.125 \\
& & Liu et al. \cite{Liu2024b} & 0.461 & 0.217 \\
& & \ourmethod & 0.543 & 0.264 \\
& & Omine+\ourmethod & \textbf{0.567} & \textbf{0.286}\\

\midrule

\multirow{5}{*}{Can. FR}
& \multirow{5}{*}{Stand-up}
& Gillick et al. \cite{Gillick2021} & 0.144 & 0.024 \\
& & Omine et al. \cite{Omine2024} & 0.237 & 0.107 \\
& & Liu et al. \cite{Liu2024b} & 0.478 & 0.170 \\
& & \ourmethod & 0.485 & 0.173 \\
& & Omine+\ourmethod & \textbf{0.521} & \textbf{0.204} \\

\midrule

\multirow{5}{*}{PT}
& \multirow{5}{*}{Stand-up}
& Gillick et al. \cite{Gillick2021} & 0.237 & 0.045 \\
& & Omine et al. \cite{Omine2024} & 0.210 & 0.057  \\
& & Liu et al. \cite{Liu2024b} & \textbf{0.402} & \textbf{0.179} \\
& & \ourmethod & 0.393 & 0.169 \\
& & Omine+\ourmethod & 0.395 & 0.167 \\

\midrule

\multirow{5}{*}{IT}
& \multirow{5}{*}{Stand-up}
& Gillick et al. \cite{Gillick2021} & 0.130 & 0.012  \\
& & Omine et al. \cite{Omine2024} & 0.391 & 0.157  \\
& & Liu et al. \cite{Liu2024b} & 0.402 & 0.195 \\
& & \ourmethod & 0.507 & \textbf{0.257} \\
& & Omine+\ourmethod & \textbf{0.545} & 0.256  \\

\midrule

\multirow{5}{*}{CS}
& \multirow{5}{*}{Stand-up}
& Gillick et al. \cite{Gillick2021} & 0.439 & 0.105 \\
& & Omine et al. \cite{Omine2024} & 0.570 & 0.272  \\
& & Liu et al. \cite{Liu2024b} & 0.438 & 0.232  \\
& & \ourmethod & 0.585 & 0.301  \\
& & Omine+\ourmethod & \textbf{0.638} & \textbf{0.321} \\

\midrule

\multirow{5}{*}{HU}
& \multirow{5}{*}{Stand-up}
& Gillick et al. \cite{Gillick2021} & 0.578 & 0.208 \\
& & Omine et al. \cite{Omine2024} & 0.706 & 0.376 \\
& & Liu et al. \cite{Liu2024b} & 0.429 & 0.281 \\
& & \ourmethod & 0.796 & \textbf{0.501}  \\
& & Omine+\ourmethod & \textbf{0.825} & 0.492  \\

\midrule

\multirow{5}{*}{RU}
& \multirow{5}{*}{Stand-up}
& Gillick et al. \cite{Gillick2021} & 0.240 & 0.066 \\
& & Omine et al. \cite{Omine2024} & 0.443 & 0.199 \\
& & Liu et al. \cite{Liu2024b} & 0.309 & 0.143  \\
& & \ourmethod & 0.438 & 0.209  \\
& & Omine+\ourmethod & \textbf{0.570} & \textbf{0.254}  \\

\bottomrule
\end{tabular}
}
\caption{Results of the models in different languages, dialects, and domains.}
\label{tab:results_full}
\end{table}

\subsection{Evaluations Metrics}

For each model, we evaluate the segmentation of laughter intervals using the Intersection over Union (IoU) metric with two thresholds: 0.3 to know how well the model detects laughter intervals, and 0.7 to evaluate the temporal segmentation. Based on these thresholds, we compute Recall and F1-score. Recall measures the proportion of ground-truth laughter segments correctly detected by the model. The F1-score is the harmonic mean of Precision and Recall. 


\subsection{Experimental Settings}

Experiments were implemented in PyTorch and scikit-learn \cite{Pedregosa2012}, and all computations were performed on an NVIDIA GeForce RTX 2080. For the audio encoder $\mathcal{A}$, we rely on the publicly available BYOL-A.\footnote{\url{https://github.com/nttcslab/byol-a}} 
The second pre-trained process was run with a batch size of 128, a learning rate of 0.0001, for 100 epochs, using a seed of 42. The Isolation Forest was configured with the contamination parameter set to \textit{auto}, which allows the model to calculate the anomaly threshold based on the intrinsic distribution of the data rather than a fixed percentage. 


\subsection{Results and Analysis}

\subsubsection{Global Results}

Table \ref{tab:results_full} reports performance across languages, dialects, and domains. Overall, while the current state of the art from Omine et al. \cite{Omine2024} remains highly competitive in US English domains, our approach demonstrates more stable and superior performance across languages and diverse conditions. 

In US English, Omine et al.'s model achieves the best results overall, in particular on Stand-up and YouTube videos, confirming its strong performance in in-domain English settings. However, F1 drops substantially on TV Shows, even when using a different probability threshold. As we discuss in Section \ref{subsec:laughter_length}, this degradation may be explained by the prevalence of longer laughter segments in TV shows, which differ from the laughter patterns observed in stand-up or online contexts.

In contrast, our method consistently outperforms competing approaches in non-English settings, achieving the best results across Spanish, French, Italian, Czech, Hungarian, and Russian. This suggests better cross-lingual robustness and generalization beyond English-centric training conditions. We believe this is due to the backbone of Omine et al.'s model, as it is a model trained for English Audio Speech Recognition (ASR) and not for Non-Semantic Speech tasks. 
Finally, it is notable that Liu et al.'s model \cite{Liu2024b} obtains the strongest results for PT, which may be related to specific recording conditions and acoustic characteristics of the dataset. 


\subsubsection{Impact of the Audio Encoder}

Table \ref{tab:encoder} reports the average performance per domain for various audio encoders. Basically, it shows that \ourmethod is not dependent on the encoder, as it performs similarly with several distinct models.  
On Stand-up, wav2clip \cite{Wu2022a} and BYOL-A \cite{Niizumi2023} perform very similarly, with BYOL-A obtaining a slightly higher F1 at IoU=0.3, while wav2clip is marginally better at IoU=0.7. On TV Shows and YouTube, BYOL-A clearly outperforms wav2clip at both overlap thresholds, suggesting that self-supervised audio representations transfer particularly well to TV show data. 

\begin{table}[h!]
\centering
\small
\begin{tabular}{c|c|ll}
\multirow{2}{*}{\textbf{Dataset}}   & \multirow{2}{*}{\textbf{Encoder}}  & \multicolumn{2}{c}{\textbf{F1}} \\
          &        & \textbf{IoU=0.3} & \textbf{IoU=0.7} \\
\hline \hline

\multirow{2}{*}{Stand-up} 
 & wav2clip \cite{Wu2022a} & 0.582 & 0.270 \\
 & BYOL-A \cite{Niizumi2023} & 0.584 & 0.269 \\
\hline

\multirow{2}{*}{TV Show} 
 & wav2clip \cite{Wu2022a} & 0.890 & 0.706 \\
 & BYOL-A \cite{Niizumi2023} & 0.910 & 0.735 \\
\hline

\multirow{2}{*}{Youtube} 
 & wav2clip \cite{Wu2022a} & 0.257 & 0.063 \\
 & BYOL-A \cite{Niizumi2023} & 0.315 & 0.087 \\
\end{tabular}
\caption{Average performance per dataset with different audio encoders using our approach.}
\label{tab:encoder}
\end{table}

\subsubsection{Laughter Length Related Performances} \label{subsec:laughter_length}

\begin{figure}[!h]
    \centering
    \includegraphics[width=\linewidth]{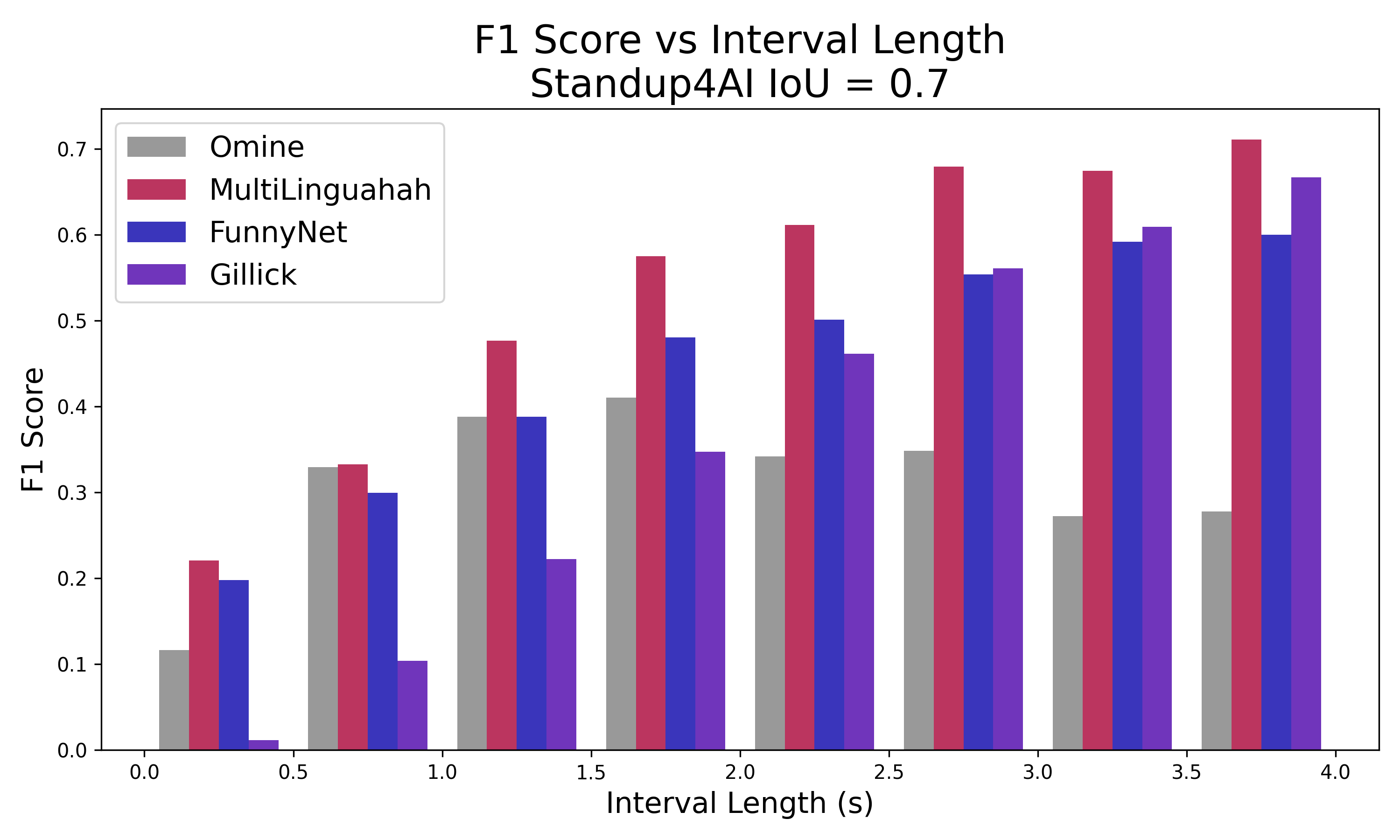}
    \caption{Comparison of F1-scores of the proposed method against three baseline models relative to laughter duration using a temporal IoU threshold of 0.7.
    }
    \label{fig:laughter_length}
\end{figure}

Figure \ref{fig:laughter_length} shows the F1 score of \ourmethod vs Omine et al.'s method on the Standup4AI dataset, with respect to the laughter duration. It is possible to see that the proposed method outperforms the baseline, in particular when the laughter events last longer.  
Once again, we believe that the pretrained ASR backbone used by Omine et al. works against it. Indeed, as soon as the laughter is too long or the speech is not that of the training set, the model is faced with Out-of-distribution data that can disrupt its predictions.  

Furthermore, our approach outperforms \cite{Liu2024b} across all laugh intervals because it better handles unpredictable noise. In noisy environments, outliers are random, which causes cluster-based strategies to fail. Because laughter shares a universal acoustic phenomenon across languages, our method succeeds by focusing on these universal patterns rather than environment-specific patterns. This allows our model to maintain precision even when the background noise is inconsistent.

\vspace*{-.2cm}
\section{Conclusion}
We introduced MultiLinguahah, an unsupervised multilingual method for acoustic laughter segmentation that frames the task as anomaly detection over energy-based segmented audio sequences. By combining a BYOL-A audio encoder with an Isolation Forest, our approach requires no labeled data and generalizes across languages and domains. 
Our experiments on four datasets covering stand-up comedy, TV shows, and YouTube content demonstrate that existing supervised methods, while competitive in US English, degrade significantly in multilingual settings. We attribute this to the ASR-oriented pretraining of models such as Omine et al., which introduces a language bias that is particularly harmful for non-English laughter. Thus, our approach is more suitable since laughter shares similar acoustic phenomena across all languages. This allows our model to focus on these universal patterns rather than environment-specific noise. In contrast, \ourmethod achieves consistent improvements in non-English languages, highlighting the value of non-semantic audio representations for this task. 
We further showed that our method is especially effective for long laughter segments, a pattern prevalent in TV shows and group laughter scenarios, and that combining \ourmethod with Omine et al.'s model yields complementary gains across several settings. 
Future work could explore adapting the energy-based segmentation to noisier environments, investigating other anomaly detection strategies, and extending the evaluation to more typologically diverse languages.  

\bibliographystyle{IEEEtran}
\bibliography{mybib,humour_standup,smila}

\end{document}